\documentclass[sigconf]{aamas}

\usepackage{balance} % for balancing columns on the final page

\usepackage{amsmath,amsfonts}
\usepackage{algorithmic}
\usepackage{graphicx}
\usepackage{textcomp}
\usepackage{xcolor}
\usepackage{balance} 
\usepackage{booktabs}
\usepackage{array}
\usepackage {subcaption}
\usepackage{siunitx}
\usepackage{makecell}
\usepackage{multirow}
\usepackage{algorithm}
\usepackage{arydshln}
\usepackage{listings}

\lstdefinestyle{aamas-tight}{
	basicstyle=\ttfamily\small,
	columns=fullflexible,
	numberstyle=\scriptsize\color{gray},
	backgroundcolor=\color{gray!12},
	frame=single,
	framerule=0.4pt,
	rulecolor=\color{black!40},    
	numbers=left,
	numbersep=6pt,
	xleftmargin=8pt,
	columns=fullflexible,
	showstringspaces=false
}
\setcopyright{none}
\acmConference{arXiv preprint}{}{}{}
\renewcommand\footnotetextcopyrightpermission[1]{}
\acmSubmissionID{<<3362>>}

\submissionType{Research Paper Track}
\title[LLM-DMGR]{Automating Domain Model Generation and Refinements for LLM-Symbolic Planning via Semantic-Enriched Agentic Framework}

\author{Xuelong Liu}
\affiliation{
	\institution{School of Artificial Intelligence, University of Chinese Academy of Sciences}
	\institution{Institute of Automation, Chinese Academy of Sciences}
	\city{Beijing}
	\country{China}}
\email{liuxuelong2022@ia.ac.cn}

\author{Jiahao Zhao}
\authornote{Corresponding authors.}
\affiliation{
	\institution{Institute of Automation, Chinese Academy of Sciences}
    \institution{Wenge AI}
    \institution{ScienceOne AI}
	\city{Beijing}
	\country{China}}
\email{zhaojiahao2019@ia.ac.cn}

\author{Qingchao Kong}
\affiliation{
	\institution{Institute of Automation, Chinese Academy of Sciences}
	\institution{School of Artificial Intelligence, University of Chinese Academy of Sciences}
	\city{Beijing}
	\country{China}}
\email{qingchao.kong@ia.ac.cn}

\author{Wenji Mao}
\authornotemark[1]
\affiliation{
	\institution{Institute of Automation, Chinese Academy of Sciences}
	\institution{School of Artificial Intelligence, University of Chinese Academy of Sciences}
	\city{Beijing}
	\country{China}}
\email{wenji.mao@ia.ac.cn}

\begin{abstract}
	Symbolic planning is a solid AI research and application field, which relies on well-defined domain models typically represented in the PDDL planning domain definition language. Automatic construction of such domain models remains a challenging task. Recent advances in large language models (LLMs) have identified the importance of symbolic planning in LLM reasoning, and motivated LLM–symbolic planning as a means to generate PDDL domain models from natural language descriptions. However, due to the semantic ambiguity and the limited self-verification capabilities of current LLMs, existing approaches mainly depend on predefined predicate sets, ground-truth domain models, or human feedback, and may produce syntactically incorrect or semantically inconsistent domain models. To address these limitations, in this paper, we propose a semantic-enriched agentic framework LLM-DMGR to automate domain model generation and refinements for LLM-symbolic planning. Our framework first extracts structured semantic knowledge from concise descriptions of domain actions to enrich LLM-based domain model generation. It then performs action-level refinement to detect syntactic errors and semantic inconsistencies in an action-by-action manner. It further applies plan-level agentic refinement, where each LLM agent utilizes  planner feedback to coordinate the refinement process via multi-agent collaboration. Experiments on ten planning domains demonstrate that our approach substantially improves problem-solving rate over representative baselines, highlighting its effectiveness in enhancing domain model generation and refinements, and LLM-based planning.
\end{abstract}


\newcommand{\BibTeX}{\rm B\kern-.05em{\sc i\kern-.025em b}\kern-.08em\TeX}

\begin{document}
	
	%%% The following commands remove the headers in your paper. For final 
	%%% papers, these will be inserted during the pagination process.
	
	%\pagestyle{fancy}
	%\fancyhead{}
	
	%%% The next command prints the information defined in the preamble.
	
	\maketitle 
	
	%%%%%%%%%%%%%%%%%%%%%%%%%%%%%%%%%%%%%%%%%%%%%%%%%%%%%%%%%%%%%%%%%%%%%%%%
	
	\section{Introduction}
	Symbolic planning \cite{ghallab2004automated} is a core research and application field in AI that centers on automated reasoning \cite{robinson2001handbook} and decision making for intelligent agents. Planning itself relies on well-defined domain models, typically represented as Planning Domain Definition Language (PDDL) \cite{mcdermott20001998}. Traditionally, constructing such PDDL domain models remains a manual process, which is time-consuming and expertise-intensive in complex domains. Moreover, maintaining syntactic correctness and semantic consistency across actions and predicates specified in domain models is another challenging undertaking. Both challenges raise the scalability and reliability issues in symbolic planning, emphasizing the urgent need to automate domain model generation and validation.
	
	With the recent advances in Large Language Models (LLMs), LLMs demonstrate  prominent capabilities in natural language understanding, generation and reasoning tasks \cite{brown2020language, ren2025deepseek,madaan2023self, qiu2025relopt}, yet they are still limited in symbolic reasoning. 
	Recent research has identified the importance of symbolic planning in LLM reasoning and proposed the LLM-symbolic planning paradigm \cite{huang2025planning}. Under this paradigm, LLMs provide a new alternative to 
	automate PDDL domain model generation \cite{huang2025planning, oswald2024large, mahdavi2024leveraging} for solving planning problems. Although promising, existing approaches to domain model generation have relied on the strong assumptions, including predefined predicate sets \cite{oswald2024large,huang2025planning}, ground-truth domain models \cite{mahdavi2024leveraging}, and human feedback \cite{guan2023leveraging}. In practice, these assumptions are difficult to satisfy, particularly in complex domains where manual specification and expert intervention are hard to obtain. Moreover, due to the semantic ambiguity and limited self-verification capabilities of current LLMs, the generated domain models frequently exhibit syntactic errors and semantic inconsistencies, affecting their reliability in planning tasks.
	
	To address the above research challenges and limitations, in this paper, we propose a semantic-enriched agentic framework to automate \textbf{D}omain \textbf{M}odel \textbf{G}eneration and \textbf{R}efinements   in \textbf{LLM}-symbolic planning (\textbf{LLM-DMGR}). Our LLM-DMGR framework consists of semantic knowledge extraction for domain knowledge enrichment and two-level feedback (i.e. action-level and plan-level) for iterative agentic refinements. 
	Specifically, semantic knowledge extraction acquires structured knowledge about domain actions and predicates from natural language descriptions through linguistic triple identification and semantic enrichment with external knowledge bases (i.e. WordNet and VerbNet). Then, LLM-DMGR introduces a bi-level feedback-driven refinement mechanism to validate the generated domain models. Action-level refinement develops a rule-based syntax validator and an LLM-based semantic evaluator to detect syntactic errors and semantic inconsistencies, focusing on the feedback of single action representation in each round of refinement. After the refinements of all actions in domain action model, plan-level agentic refinement leverages LLM agents to evaluate the representation errors and planning failures on the generated domain model, by function-calling symbolic planners as external tools. The evaluation of each agent is summarized via multi-agent collaboration to form plan-level feedback for the refinement process. Experimental results on ten representative planning domains from the International Planning Competition (IPC) generally verify the efficacy of our proposed framework. 
	
	\begin{figure*}[h]  
		\centering
		\includegraphics[width=0.95\linewidth]{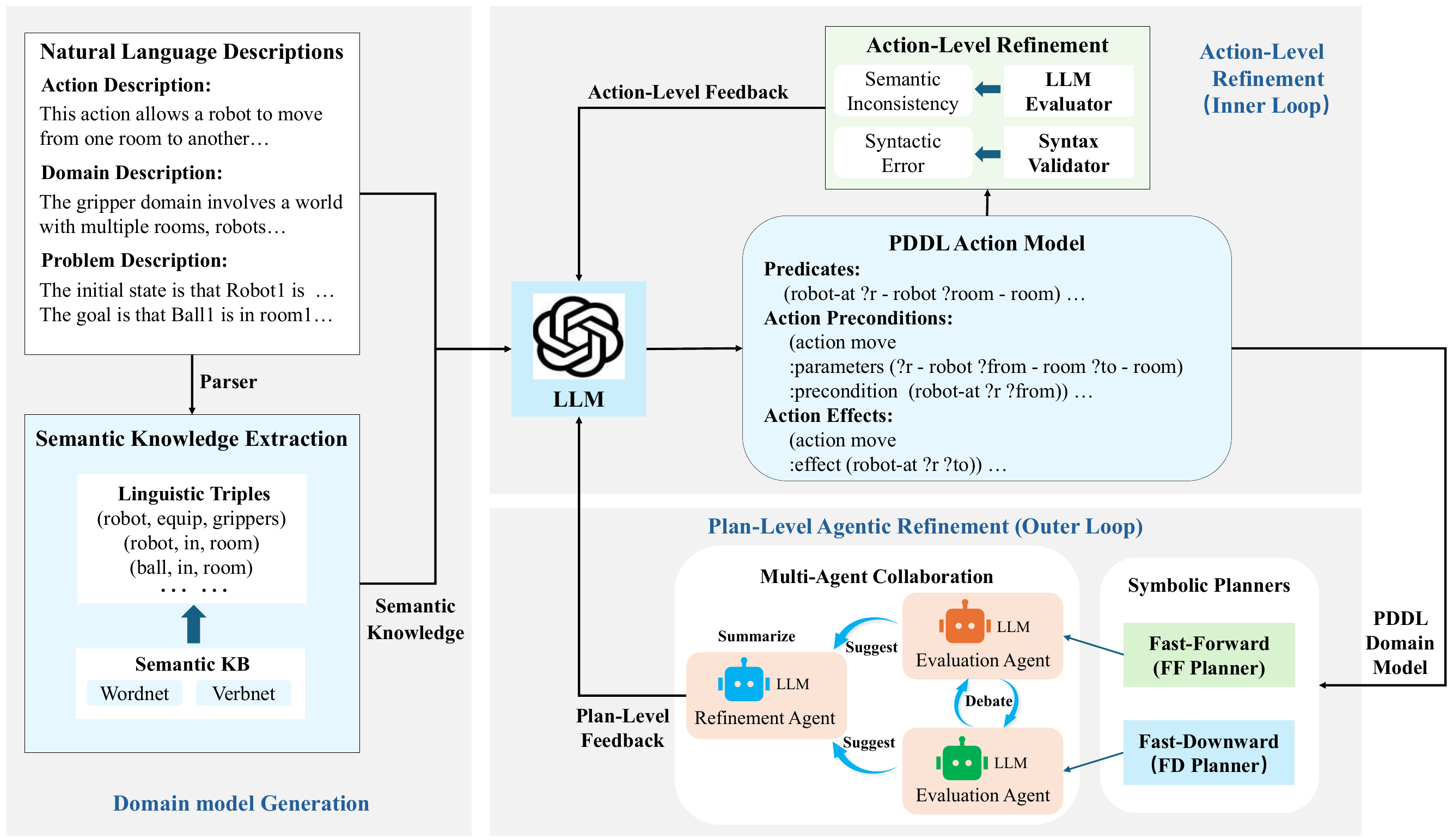}
		\caption{Overall structure of our semantic-enriched agentic framework LLM-DMGR for automated domain model generation and refinements.}
		\label{fig:Our Framework}
	\end{figure*}

	The main contributions of our work are as follows:
	\begin{itemize}
		%\item We propose \textbf{LLM-DMGR}, a semantic-enriched agentic framework for automated domain model generation and refinements for LLM-symbolic planning, which is enhanced by semantic knowledge extraction and agentic refinements at both action and plan levels.
		
		\item We propose \textbf{LLM-DMGR}, a semantic-enriched agentic framework for automated domain model generation and refinements for LLM-symbolic planning, which performs semantic knowledge extraction and agentic refinements at both action and plan levels.
		
		\item To eliminate the reliance on predefined predicates, we develop semantic knowledge extraction and action-level refinement in LLM-DMGR, which utilize linguistic patterns enriched with external semantic knowledge bases.

		%\item To avoid the reliance on human intervention or domain ground-truth, the proposed plan-level agentic refinement in LLM-DMGR incorporates an external planner for individual LLM agent, and refines plan-level feedback through multi-agent collaboration.  
		
		\item To avoid the reliance on human intervention or domain ground-truth, we develop plan-level agentic refinement in LLM-DMGR, which incorporates external planners for LLM agents and refines plan feedback through multi-agent collaboration.  
		
		\item We conduct extensive experiments across ten planning domains, and the results demonstrate the effectiveness of our framework in enhancing domain model generation and refinements, and LLM-based planning.

		%enhancing LLM planning capability as well as the efficacy of domain model generation.
		
	\end{itemize}

	%%%%%%%%%%%%%%%%%%%%%%%%%%%%%%%%%%%%%%%%%%%%%%%%%%%%%%%%%%%%%%%%%%%%%%%%
	\section{Related Works}
	
	Existing studies on LLM-based planning are conducted mainly from two perspectives: LLMs as planners and LLM–symbolic planning.
	
	\subsection{LLMs as Planners}
	
	%LLMs exhibit strong reasoning and in-context learning capabilities across mathematics~\cite{chen2025seed, ren2025deepseek}, programming~\cite{qiu2025relopt,ni2023lever}, and natural language processing tasks~\cite{brown2020language}. On this basis, 
	
	The recent advances in LLMs have witnessed LLMs' strong  planning and reasoning capabilities in tasks such as mathematics, programming and natural language processing. Most existing research in LLM reasoning develops task-oriented approaches without particular focus on symbolic planning, with the exception of LLM as planners. Several works have been proposed to treat LLMs as planners, which directly generate action sequences from textual task descriptions~\cite{wei2022chain, shen2023hugginggpt}. However, due to the lack of feedback, they are prone to error accumulation. To alleviate this, several works on LLMs as planners ~\cite{yao2023react, huang2023inner} propose to  iteratively refine plans using external feedback, focusing on short-horizon decisions and overlook long-term goal structure. Thus, to facilitate long-horizon reasoning, several works~\cite{yao2023tree, zhao2023large, meng2024llm} further integrate LLMs with search methods, where LLMs are typically used as heuristics or evaluators to estimate action/state quality. Despite their effectiveness, these search-based approaches often require substantial computational cost, limiting their scalability in complex domains. 
	
	\subsection{LLM-Symbolic Planning}
	
	Symbolic planning as an AI arena typically requires automated reasoning techniques to solve complex planning problems for decision-making agents. With the advance of LLM reasoning, recent research proposes the LLM-symbolic planning paradigm to automate domain model and problem generation for LLM-based planning.
	Several works~\cite{liu2023llm+, xie2023translating} translate natural language into PDDL problem instances while assuming predefined domain models. 
	Since domain model generation is typically more challenging than problem generation, recent works have focused on automated domain model generation from natural language using LLMs. Oswald et al.~\cite{oswald2024large} focus on generating domain models from predefined predicate sets, whereas Mahdavi et al.~\cite{mahdavi2024leveraging} propose an exploration-walk metric for iterative refinement assuming access to ground-truth domain models. Huang et al.~\cite{huang2025planning} evaluate the semantic consistency between generated domain models and natural-language descriptions using a trained sentence encoder, while filtering irrelevant outputs based on predefined predicate sets. Guan et al.~\cite{guan2023leveraging} generate PDDL domain models with human feedback. %while \textsc{NL2Plan}~\cite{gestrin2024nl2plan} generates both domain and problem models from natural language	and supports feedback from humans or another LLM. %In general, existing approaches still depend on predefined predicate sets, ground-truth-dependent signals, or human feedback, which limits their applicability in fully automated NL-to-domain settings.
	Besides, the semantic ambiguity of natural language and the limited self-verification capabilities of current LLMs also bring about semantic inconsistency issues in these approaches.
	In contrast to these lines of work, our experiments focus on matched-input baselines and a reproduced P\&D-Tree pipeline, and we summarize the assumption differences in Table~\ref{tab:cmp_assume}.
	%difficulty to maintain semantic consistency of domain models.
	
	\iffalse
	\begin{figure*}[t]
		\centering
		\includegraphics[width=0.9\linewidth]{Pictures/PDDL-Domain-1}
		%\vspace{0.05cm}
		\includegraphics[width=0.9\linewidth]{Pictures/PDDL-Problem-1}
		\caption{PDDL domain representation for Grippers, and examples of Grippers problem representation and plan solution to the problem.}
		\label{fig:pddl_example}
	\end{figure*}
	\fi

	%%%%%%%%%%%%%%%%%%%%%%%%%%%%%%%%%%%%%%%%%%%%%%%%%%%%%%%%%%%%%%%%%%%%%%%%
	\section{Task Formalization and Representation}
	We first introduce the PDDL representation, and then present the formalization of our framework.
	
	\subsection{PDDL Representation}
	
	The PDDL domain model is denoted as $\mathbb{D} = \langle P, A \rangle$, where $P$ is the set of predicates describing the essential properties of the environment, and $A$ is the action set that can be performed within the domain. Each action specifies a set of parameters it manipulates, preconditions that must be satisfied before execution and effects that are produced after execution. In essence, the PDDL domain model captures the transition dynamics of the environment.
	In addition, a PDDL problem is denoted as $\mathbb{P} = \langle O, I, G \rangle$, where $O$ is a finite set of objects grounding the domain $\mathbb{D}$, $I$ denotes the initial state and $G$ denotes the goal state of the planning problem. The PDDL problem representation instantiates the PDDL domain model to form a concrete instance of a planning environment. 
	The solution to a PDDL planning problem $\mathbb{P}$ is a sequence of actions $\pi = (a_0, a_1, \ldots, a_n)$ that transforms the initial state $I$ into the goal state $G$. 
	
	% Figure \ref{fig:pddl_example} illustrates the PDDL domain representation for Grippers, an example of PDDL Grippers problem representation and an example of PDDL plan solution to Grippers problem.

	\subsection{Domain Model Generation and Refinement}
	Different from the \textit{LLMs as Planners} paradigm formalized in (\ref{eq_direct_planning}), which leverages the LLM to directly generate a plan from natural language descriptions of a PDDL domain $\mathbb{NL}(\mathbb{D})$ and a PDDL problem $\mathbb{NL}(\mathbb{P})$, we take the LLM-Symbolic Planning paradigm.
	\begin{equation}
		\pi \sim P_{LLM}(\cdot \mid \mathbb{NL}(\mathbb{D}), \mathbb{NL}(\mathbb{P})).
		\label{eq_direct_planning}
	\end{equation}
	Existing LLM-symbolic planning approaches have relied on predefined predicate sets, ground-truth domain models or human feedback for domain model generation. In contrast, we develop the PDDL domain model generation and bi-level refinement processes in our LLM-DMGR approach, formalized in (\ref{eq_ling_modeler_refine}). Specifically, to eliminate the reliance on predefined predicates, our work proposes semantic knowledge based extraction $\mathbb{SK}_{\mathbb{D}}$ (Note that compared to previous LLM-symbolic planning research, our approach utilizes more concise action descriptions $\mathbb{NL}(\mathbb{D}-)$, without explicitly specifying action preconditions and effects) and action-level refinement $\mathcal{F}_{action}$. In addition, to avoid the reliance on human feedback or domain ground-truth, our work proposes plan-based agentic refinement $\mathcal{F}_{plan}^{MAC}$ on top of the action-level refinement, based on multi-agent collaboration (\textit{MAC}). For each domain, we use a single problem $\mathbb{P}_{\mathrm{gen}}$ to generate and refine the domain model. During evaluation, the resulting domain model is kept fixed and tested on ten problems $\mathbb{P}_{\mathrm{test}}$ that are used exclusively for evaluation.
	\begin{equation}
		\label{eq_ling_modeler_refine}
		\begin{split}
			\mathbb{SK}_{\mathbb{D}} &= \mathbb{SK}(\mathbb{NL}(\mathbb{D-}), \mathbb{NL}(\mathbb{P}_{\mathrm{gen}}), \\
			\mathbb{D}_{generate} &\sim P_{LLM}(\cdot \mid \mathbb{NL}(\mathbb{D-}), \mathbb{NL}(\mathbb{P}_{\mathrm{gen}}), \mathbb{SK}_{\mathbb{D}}), \\
			\mathbb{D}_{refine} & \sim P_{LLM}(\cdot \mid\mathbb{D}_{generate}, \mathcal{F}_{action}, \mathcal{F}_{plan}^{MAC}), \\
			\pi &= \operatorname{Planner}(\mathbb{D}_{refine}, \mathbb{P}_{\mathrm{test}}).
		\end{split}
	\end{equation}
	%%%%%%%%%%%%%%%%%%%%%%%%%%%%%%%%%%%%%%%%%%%%%%%%%%%%%%%%%%%%%%%%%%%%%%%%
	\section{Proposed Framework}
	
	%In this section, we present LLM-DMGR, a semantic-enriched agentic refinement framework which, as shown in Figure~\ref{fig:Our Framework}, consists of semantic knowledge extraction for domain knowledge enrichment, action-level refinement and plan-level agentic refinement for iterative refinements.
	
	In this section, we present LLM-DMGR, a semantic-enriched agentic framework form domain model generation and refinements. As illustrated in Figure~\ref{fig:Our Framework}, the framework consists of three key components: semantic knowledge extraction, action-level refinement, and plan-level agentic refinement, which can improve the quality of generated domain models through iterative refinement.

	\subsection{Semantic Knowledge Extraction}
	
	In our approach, semantic knowledge extraction is conducted through a two-stage pipeline: linguistic triple identification based on syntactic dependency parsing, and semantic knowledge enrichment via external knowledge bases.
	
	\subsubsection{Linguistic Triples Identification}

	In PDDL domain models, predicates encode entity properties and relations, which are often reflected in natural language through adjectives, verbs, and prepositional phrases. To systematically capture such information, we employ the Stanza parser for part-of-speech tagging and dependency parsing, enabling the identification of entity properties from adjectival modifiers and relations from verbal and prepositional structures. Based on the resulting dependency structures, we construct a set of linguistic triples, e.g. (robot, equip, gripper) and (robot, at, room), to represent the underlying semantics of the domain. Specifically, verbs are treated as relation triggers for extracting subject–relation–object triples, prepositions are normalized into semantic relations, and passive constructions as well as nominal modifiers are handled to improve coverage. Redundant or ill-formed triples are further filtered to enhance correctness. Through this process, the identified triples encode semantic relations and implicit parameter information, thereby providing semantic guidance for LLMs to generate accurate and consistent domain model.

	\subsubsection{Semantic Knowledge Enrichment}
	\begin{figure}[t]  
		\centering
		\includegraphics[width=0.9\linewidth]{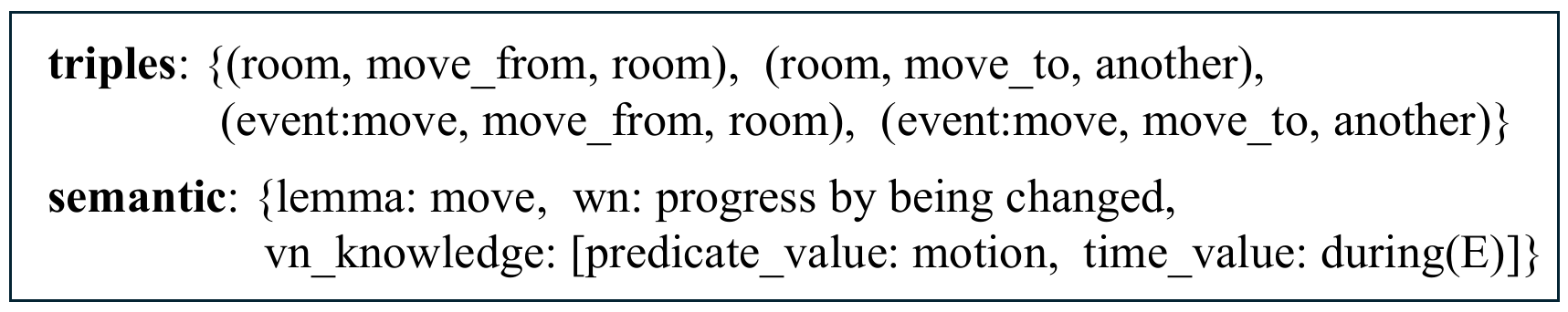}
		\caption{An example of extracted semantic knowledge for Grippers.}
		\label{fig:Semantic}
	\end{figure}
	To enrich the semantic information associated with the extracted triples, we further incorporate external semantic knowledge from WordNet and VerbNet. WordNet provides lexical definitions of concepts, while VerbNet offers structured semantic descriptions of verbs, including their argument structures and semantic roles. For nouns and verbs appearing in the triples, we select the most relevant WordNet synsets by comparing the natural language context with candidate lexical definitions. For each verb, we further retrieve VerbNet knowledge by selecting the highest-ranked semantic frames based on their semantic similarity to the selected WordNet definitions, thereby enriching the linguistic triples with richer action semantics. Through this process, the enriched linguistic triples provide richer semantic guidance for subsequent predicate and action generation.
	
	Figure \ref{fig:Semantic} presents an example of the extracted semantic knowledge for the Grippers domain.
	
	\iffalse
	\begin{lstlisting}[style=aamas-tight, gobble=7,
		caption={An example of extracted semantic knowledge for Grippers.}, label={lst:grippers-SK}]
		'triples':
		[('room', 'move_from', 'room'), ('room', 'move_to', 'another'), 
		('event:move', 'move_from', 'room'), 
		('event:move', 'move_to', 'another')]
		
		'verb_semantic': {'lemma': 'move', 'wn': 'progress by being changed',
			'vn_knowledge': [{'predicate_value': 'motion', 
				'time_value': 'during(E)'}]}
		
	\end{lstlisting}
	\fi

	\subsection{Action-Level Refinement}
	
	Action-level refinement aims to improve the syntactic correctness and semantic consistency of generated action representations through iterative feedback. Specifically, it employs rule-based syntax validation and LLM-based semantic evaluation to detect syntactic errors and semantic inconsistencies in an action-by-action manner, enabling targeted refinement of individual action in each iteration. In our framework, the feedback captures common issues such as invalid parameter types, invalid predicate definitions, incorrect predicate usage, duplicated or inconsistent predicates, and  inconsistent parameter ordering. Table~\ref{tab:error_analysis_act} presents representative examples of the action-level feedback used in our framework.

	The refinement process first employs a syntax validator to examine whether each generated action representation conforms to the required PDDL format. Specifically, the syntax validator checks parameter declarations, object types, predicate definitions, predicate names, and predicate usage in preconditions and effects, while also detecting unsupported constructs and malformed outputs. After syntactic validation, semantic evaluation is performed to identify inconsistencies in the generated predicates and action representations.  For example, the semantic evaluator identifies duplicated or semantically inconsistent predicates, encourages the reuse of more general predicate forms, and verifies the consistency of parameter ordering in relational predicates.

	The detected errors and inconsistencies are then transformed into refinement feedback and returned to the LLM for revision. The feedback provides concrete guidance, such as correcting invalid parameter types, removing unsupported constructs, reusing existing predicates instead of introducing redundant ones, and adjusting incorrect argument ordering. Through iterative refinement, LLM-DMGR progressively improves the syntactic correctness and semantic consistency of the generated action representations.

	\begin{table*}[t]
		\centering
		%\small
		%\footnotesize
		%\scriptsize
		\setlength{\tabcolsep}{4pt}
		\renewcommand{\arraystretch}{1.1}
		\caption{Illustrations of action-level feedback}
		\label{tab:error_analysis_act}
		%\resizebox{\textwidth}{!}{
			\begin{tabular}{
					>{\raggedright\arraybackslash}p{2.5cm} 
					>{\raggedright\arraybackslash}p{5cm}
					>{\raggedright\arraybackslash}p{9cm}
				}
				
				\toprule
				\textbf{Feedback Type} 
				& \textbf{Inconsistency/Error} 
				& \textbf{Example} \\
				\midrule
				
				% ================= Syntactic Error =================
				\multirow{3}{*}{
					\begin{tabular}{@{}l@{}}
						\textbf{Syntactic} \\
						\textbf{Error}
					\end{tabular}
				}
				& Use of undefined PDDL types
				& \textit{(at ?obj - cargo ?loc - location)}, where \textit{cargo} is undefined. \\
				\cdashline{2-3}
				
				& Unsupported predicate format
				& When using \textit{(= ?x ?y)}, \textit{(:equality)} is not declared. \\
				\cdashline{2-3}
				
				& Invalid predicate usage
				& One parameter \textit{robot} is missing in \textit{(gripper-free gripper1)}. \\
				
				\cline{1-3}
				
				% ================= Semantic Inconsistency =================
				\multirow{3}{*}{
					\begin{tabular}{@{}l@{}}
						\textbf{Semantic} \\
						\textbf{Inconsistency}
					\end{tabular}
				}
				& Duplicated predicate generation
				& \textit{(tile-clear ?t - tile)} and \textit{(tile-free ?t - tile)} have the same semantic meaning but are generated as different predicates. \\
				\cdashline{2-3}
				
				& Inconsistent predicate generation
				& \textit{(tray\_at\_kitchen ?t - tray)} and \textit{(tray\_at ?t - tray ?l - location)} have similar semantic meanings but use different predicate forms. \\
				\cdashline{2-3}
				
				& Misorder of parameters in predicates
				& \textit{(walk ?to ?from)}, where \textit{?from} and \textit{?to} are misordered. \\
				
				\bottomrule
			\end{tabular}
			%}
	\end{table*}

	\begin{table*}[t]
		\centering
		%\small
		%\scriptsize
		\setlength{\tabcolsep}{4pt}
		\renewcommand{\arraystretch}{1.1}
		\caption{Illustrations of plan-level feedback}
		\label{tab:error_analysis_plan}
		%\resizebox{\textwidth}{!}{
			\begin{tabular}{
					>{\raggedright\arraybackslash}p{2.3cm} 
					>{\raggedright\arraybackslash}p{5.6cm}
					>{\raggedright\arraybackslash}p{9cm}
				}
				
				\toprule
				\textbf{Feedback Type} 
				& \textbf{Inconsistency/Error} 
				& \textbf{Example} \\
				\midrule
				
				\multirow{2}{*}{
					\begin{tabular}{@{}l@{}}
						\textbf{Representation} \\
						\textbf{Error}
					\end{tabular}
				}
				& Use of unsupported feature
				& Using unsupported keywords, such as \textit{forall} or \textit{exists}, in the domain model representation. \\
				\cdashline{2-3}
				
				& Invalid predicate definition
				& Syntactic errors in predicate definitions. \\
				
				\cline{1-3}
				
				\multirow{3}{*}{
					\begin{tabular}{@{}l@{}}
						\textbf{Planning} \\
						\textbf{Failure}
					\end{tabular}
				}
				& Search stopped without finding a solution
				& The planner reports error information without finding a valid plan. \\
				\cdashline{2-3}
				
				& Time limit has been reached
				& The planner reports intermediate information without returning a plan within the time limit. \\
				\cdashline{2-3}
				
				& Unreachable goal state
				& Incomplete causal dependencies between actions. Missing predicates or effects should be added to establish valid causal transitions. \\
				
				\bottomrule
			\end{tabular}
			%}
	\end{table*}

	\begin{table*}[t]
		\centering
		%\small
		%\scriptsize
		%\tiny
		\setlength{\tabcolsep}{4pt}
		\renewcommand{\arraystretch}{0.9}
		\caption{The information of the ten evaluated domains}
		\label{tab:domain_info}
		%\resizebox{\textwidth}{!}{
			\begin{tabular}{ccccccccccc}
				\toprule
				\textbf{Domain} 
				& Grippers 
				& Grippers-ood 
				& Miconic 
				& Movie 
				& Driverlog
				& Hiking 
				& Termes 
				& Floortile
				& Childsnack 
				& Barman \\
				\midrule
				\textbf{\#Precondition + \#Effect} 
				%\makecell{\textbf{\#Precond}\\\textbf{\#Effect} }
				& 14 
				& 18 
				& 16 
				& 16 
				& 28 
				& 64 
				& 47
				& 51 
				& 37 
				& 109 \\
				\textbf{Max.\#Object}
				%\makecell{\textbf{Max.} \\\textbf{\#Object}}
				& 20 
				& 20 
				& 25 
				& 115 
				& 26 
				& 20 
				& 25 
				& 24 
				& 27 
				& 21 \\
				\bottomrule
			\end{tabular}
			%}
	\end{table*}

	\subsection{Plan-Level Agentic Refinement}
	
	%After all actions are generated, they are aggregated into a complete PDDL domain model, which is then evaluated by attempting to solve the corresponding PDDL problems. Instead of relying on a single planner, we introduce a plan-level agentic refinement process in which multiple classical planners are used to test the generated domain model and provide planning feedback, including unreachable goal states, search timeouts, and memory exhaustion. If at least one planner finds a valid plan, the current domain model is considered valid and the generation process terminates. Otherwise, the feedback from different planners is collected and assigned to multiple LLM-based refinement agents for further analysis and refinement.
	
	After all actions are generated, they are aggregated into a complete PDDL domain model. The LLM converts the natural-language problem description $\mathbb{NL}(\mathbb{P}_{\mathrm{gen}})$ into the PDDL problem $\mathbb{P}_{\mathrm{gen}}$ using the predicates defined in the generated domain. Multiple classical planners evaluate the current model on the same generation problem $\mathbb{P}_{\mathrm{gen}}$ and provide feedback, including unreachable goals, search timeouts, and memory exhaustion. The generation problem remains unchanged throughout the refinement process. If at least one planner finds a plan for this problem, the planning check is passed and generation terminates. Otherwise, feedback from the planners is assigned to multiple LLM-based refinement agents for further analysis and model revision.
	
	%After all actions are generated, they are assembled into a complete PDDL domain model. At each refinement iteration, the LLM constructs a PDDL representation of the same generation task using the predicates defined in the current domain model. The underlying task remains fixed, while its PDDL representation is updated to accommodate changes in the domain model. Multiple classical planners evaluate the current domain–problem pair and provide feedback, including representation errors, unreachable goals, search timeouts, and memory exhaustion. If at least one planner finds a plan, the planning feasibility check is passed and generation terminates. Otherwise, the planner feedback is passed to the refinement agents for further analysis and domain revision. Passing this check establishes feasibility under the current representation; it does not guarantee that the domain model fully captures the intended task semantics.

	To improve refinement reliability, we employ a multi-agent collaboration mechanism at the plan level. Specifically, feedback from different symbolic planners (e.g. FD and FF) is first analyzed by two evaluation agents, each generating planner-specific revision suggestions. Then the two agents conduct a debate process to compare and assess their suggestions. Based on the debated results, a refinement agent summarizes and synthesizes the most effective revision suggestions into plan-level feedback. The plan-level feedback is subsequently used to guide the LLM in revising the generated PDDL domain model, thereby iteratively improving the syntactic/semantic consistency at the plan level.
	
	Table~\ref{tab:error_analysis_plan} presents representative examples of the feedback used during plan-level agentic refinement. These feedback signals characterize common representation errors and planning failures, including unsupported features, invalid predicate definitions, unreachable goals, search timeouts, and memory exhaustion. Unlike action-level refinement, which focuses on the local syntactic correctness and semantic consistency of individual action, plan-level refinement evaluates whether the complete domain model can support successful planning. Therefore, it provides global feedback that complements action-level refinement by exposing modeling defects that only emerge during planning.

	\subsection{Algorithm for LLM-DMGR Framework}
	\begin{algorithm}[!t]
		\caption{Algorithm for LLM-DMGR}
		\label{alg:algorithm}
		\textbf{Input:} Concise domain description $\mathbb{NL}(\mathbb{D}-)$;\\
		\qquad generation problem description $\mathbb{NL}(\mathbb{P}_{\mathrm{gen}})$; action set $A$\\
		\textbf{Output:} Refined PDDL domain $\mathbb{D}_{ref}$\\
		\textbf{Hyperparameters:} $T_{act}=3$, $T_{plan}=3$\\
		\begin{algorithmic}[1]
			\STATE $\mathbb{SK} \gets$ Semantic-Knowledge-Extraction$(\mathbb{NL}(\mathbb{D}-),\mathbb{NL}(\mathbb{P}_{\mathrm{gen}}))$
			\STATE $P \gets \varnothing$;\; $\mathcal{F}_{plan}^{MAC} \gets \varnothing$
			\FOR{$t = 1$ to $T_{plan}$}
			\STATE $Act_{ref} \gets \varnothing$
			\FOR{each action $a \in A$}
			\STATE $\mathcal{F}_{action}^{a} \gets \varnothing$
			\FOR{$k = 1$ to $T_{act}$}
			\STATE $(Act^{a}_{gen}, P_{gen}) \gets P_{LLM}(\mathbb{SK},\mathbb{NL}(\mathbb{D}-),P,\mathcal{F}_{action}^{a},\mathcal{F}_{plan}^{MAC})$
			
			\STATE $\mathcal{F}_{action}^{a} \gets$ Action-Level-Refinement$(Act^{a}_{gen},P_{gen})$
			\STATE $Act^{a}_{ref} \gets Act^{a}_{gen}$
			\IF{$\mathcal{F}_{action}^{a} = \varnothing$}
			\STATE $P \gets P \cup P_{gen}$
			\STATE \textbf{break}
			\ENDIF
			\ENDFOR
			\STATE $P \gets P \cup P_{gen}$;\; $Act_{ref} \gets Act_{ref} \cup \{Act^{a}_{ref}\}$
			\ENDFOR
			\STATE $\mathbb{D}_{ref} \gets$ Assemble-Domain$(Act_{ref},P)$
			\STATE $\mathbb{P}_{\mathrm{gen}} \gets$
			$P_{LLM}$$(\mathbb{NL}(\mathbb{P}_{\mathrm{gen}}),
			\mathbb{D}_{ref})$
			\STATE $Plans \gets$ Multi-Planners$(\mathbb{D}_{ref},\mathbb{P}_{\mathrm{gen}})$ %\COMMENT{e.g., FD and FF}
			%\IF{at least one plan in $Plans$ is valid}
			\IF{at least one planner returns a plan for $\mathbb{P}_{\mathrm{gen}}$}
			\STATE \textbf{return} $\mathbb{D}_{ref}$
			\ELSE
			\STATE $\mathcal{F}_{plan}^{MAC} \gets$ Plan-Level-Agentic-Refinement$(Plans)$
			%\COMMENT{evaluate $\rightarrow$ debate $\rightarrow$ synthesize feedback}
			\ENDIF
			\ENDFOR
			\STATE \textbf{return} $\mathbb{D}_{ref}$
		\end{algorithmic}
	\end{algorithm}
	
	As illustrated in Algorithm~\ref{alg:algorithm}, LLM-DMGR first performs semantic knowledge extraction once to construct structured domain knowledge from natural language descriptions.
	Based on the extracted knowledge, it incrementally generates predicates and PDDL action representations, including action parameters, preconditions, and effects.
	The generated predicates are maintained in a dynamic predicate set $P$ and reused across different actions to ensure consistency.
	
	Each generated action is then validated by action-level refinement, which detects syntactic errors and semantic inconsistencies and provides action-level feedback for iterative correction within at most $T_{act}$ rounds.
	After all actions are refined, they are aggregated into a complete PDDL domain model.
	The current domain model is evaluated on the same generation problem $\mathbb{P}_{\mathrm{gen}}$ using multiple classical planners (e.g., FD and FF).
	If planning fails, planner feedback is analyzed by plan-level agentic refinement through multi-agent collaboration (evaluate, debate, and synthesize) to produce $\mathcal{F}_{plan}^{MAC}$, which guides the next outer iteration (at most $T_{plan}$ rounds).

	Through the interaction of semantic knowledge extraction, action-level refinement, and plan-level agentic refinement, LLM-DMGR progressively improves the syntactic correctness and semantic consistency of generated domain models. The prompts used in LLM-DMGR are provided in the Appendix.
	
	%%%%%%%%%%%%%%%%%%%%%%%%%%%%%%%%%%%%%%%%%%%%%%%%%%%%%%%%%%%%%%%%%%%%%%%%
	
	\section{Experiments}
	
	In this section, we conduct comprehensive experiments to evaluate the performance of our proposed  LLM-DMGR for domain model generation.

	\subsection{Domains and Data Preparation}
	We conduct extensive experiments on ten classical planning domains from the International Planning Competition benchmarks, all of which are provided in PDDL format, with ten evaluation planning problems per domain, following the experimental setup of the prior work ~\cite{mahdavi2024leveraging}. These evaluation problems are converted into PDDL to match the generated domain model, while the domain model itself remains fixed during evaluation. We run each experiment independently three times and report the average results.
	
	Table~\ref{tab:domain_info} summarizes the ten evaluated planning domains, including the total number of action preconditions and effects, and the maximum number of objects, which together reflect the complexity of domain model generation and problem difficulty in our experimental evaluation. Accordingly, Grippers, Grippers-ood, Miconic, Movie, and Driverlog are considered as simpler domains, whereas Hiking, Termes, Floortile, Childsnack, and Barman are regarded as more complex domains.

	For the natural language descriptions used in our approach, we follow the procedure of  the prior work~\cite{mahdavi2024leveraging}, which back-translates PDDL domain models into natural language using LLMs and then performs manual verification to ensure correctness. Domain descriptions provide an overview of the planning domain, whereas problem descriptions specify the initial state and goal conditions of a particular planning instance, without relying on predefined predicate sets or access to the original PDDL problem files. Unlike the prior work \cite{mahdavi2024leveraging}, which employs detailed action descriptions that explicitly enumerate action preconditions and effects, our action descriptions only describe the state changes before and after action execution. Note that, as in the prior work \cite{mahdavi2024leveraging}, we also provide the action name and parameters with their corresponding types when generating PDDL domain models. 
	
	To better illustrate the natural language descriptions we used in our experiments, we also present the \textit{Miconic} domain as an example in Section~1 of the Appendix.

	\subsection{Baselines and Evaluation Metric}
	
	\begin{table}[t]  
		\centering  
		%\scriptsize  
		\setlength{\tabcolsep}{1pt}  
		\renewcommand{\arraystretch}{1}  
		\caption{Qualitative comparison of domain-model generation methods.}  
		\label{tab:cmp_assume}  
		
		\begin{tabular}{lcccc}  
			\toprule  
			\textbf{Method}  
			& \makecell{Predicate}  
			& \makecell{Ground-\\truth}  
			& \makecell{Human\\intervention}  
			& Output \\  
			\midrule  
			
			Oswald et al.~\cite{oswald2024large}  
			& Yes & No & No & Domain \\  
			
			Huang et al.~\cite{huang2025planning}  
			& Yes & No & No & Domain \\  
			
			Guan et al.~\cite{guan2023leveraging}  
			& No & No & Yes & Domain \\  
			
			%NL2Plan~\cite{gestrin2024nl2plan}  
			%& No & No & Yes & Domain+Prob. \\  
			
			P\&D-Tree~\cite{mahdavi2024leveraging}  
			& No & Yes & No & Domain+Prob. \\  
			
			%\makecell[l]{Direct / CoT / SC-CoT /\\ Self-Refine}  
			%& No & No & No & Domain \\  
			
			LLM-DMGR (Ours)  
			& No & No & No & Domain \\  
			
			\bottomrule  
		\end{tabular}  
	\end{table}

	Prior domain-generation systems often rely on stronger external assumptions than our setting, such as predefined predicate sets~\cite{oswald2024large, huang2025planning}, ground-truth domain models~\cite{mahdavi2024leveraging}, or human intervention~\cite{guan2023leveraging}.
	Table~\ref{tab:cmp_assume} summarizes these differences against LLM-DMGR.
	%Because such mismatched assumptions make an apples-to-apples reimplementation difficult for every prior system, we adopt a two-part comparison strategy.
	Because prior methods rely on different assumptions and input settings, it is difficult to reimplement all of them under exactly the same experimental conditions. We therefore adopt a two-part comparison strategy.
	First, under the \emph{same input setting} as LLM-DMGR (natural-language domain/action descriptions and domain templates, without predefined predicates or human correction), we evaluate strong controllable baselines: (i) two \emph{direct} LLM generators in a single pass (DeepSeek-v4-flash and GPT-5-mini); and (ii) stronger matched-input baselines on DeepSeek-v4-flash and GPT-5-mini, namely Chain-of-Thought (CoT)~\cite{wei2022chain}, Self-Consistency CoT (CoT-SC)~\cite{wang2023selfconsistency}, and Self-Refine~\cite{madaan2023self}. Self-consistency CoT samples three CoT generations and selects the final domain by majority voting over normalized PDDL texts. Self-Refine first drafts a domain in the same manner as CoT, then iteratively generates feedback on the current domain and revises it for three rounds.
	%Second, we further report a same-backbone reproduction of the P\&D-Tree variant of Mahdavi et al.~\cite{mahdavi2024leveraging} with DeepSeek-v4-flash as a competitive prior method that uses environment interaction for iterative refinement.
	Second, we further report same-backbone reproductions of the P\&D-Tree variant of Mahdavi et al.~\cite{mahdavi2024leveraging} with DeepSeek-v4-flash and GPT-5-mini as a competitive prior method that uses ground-truth domain models for iterative refinement.
	LLM-DMGR is evaluated with DeepSeek-v4-flash and GPT-5-mini.
	Together, these experiments compare matched-input baselines and a reproduced prior refinement pipeline. Table~\ref{tab:cmp_assume} provides an additional qualitative comparison with other domain-model generation methods. %Together, these comparisons cover both matched-input baselines and a representative prior refinement pipeline, while Table~\ref{tab:cmp_assume} clarifies why systems that require human feedback or predefined predicates are retained as qualitative references rather than forced into an unfair quantitative comparison.

	\begin{table*}[t]
		\centering
		\setlength{\tabcolsep}{2.5pt}
		\renewcommand{\arraystretch}{1.0}
		\caption{Problem-solving rates across ten symbolic planning domains. }
		\label{tab:main}
		
		\begin{tabular}{@{}ll*{11}{c}@{}}
			\toprule
			\textbf{Backbone}
			& \textbf{Method}
			& Grippers
			& Grippers-ood
			& Miconic
			& Movie
			& Driverlog
			& Hiking
			& Termes
			& Floortile
			& Childsnack
			& Barman
			& \textbf{Avg.} \\
			\midrule
			
			%\multirow{6}{*}{\makecell{\footnotesize DeepSeek\\\footnotesize v4-flash}}  \makecell{\footnotesize Direct LLM\\\footnotesize (Single-pass)}
			\multirow{6}{*}{\makecell{DeepSeek\\-v4-flash}}
			&   \makecell[l]{Direct LLM\\(Single-pass)}
			& 1.00 & 1.00 & 0.73 & 1.00 & 0.10
			& 0.83 & 0.33 & 0.27 & 0.00 & 0.00 & 0.53 \\
			
			& CoT
			& 1.00 & 1.00 & 1.00 & 1.00 & 0.97
			& 1.00 & 0.67 & 0.97 & 1.00 & 0.00 & 0.86 \\
			
			& CoT-SC
			& 1.00 & 1.00 & 1.00 & 1.00 & 1.00
			& 1.00 & 0.90 & 0.97 & 1.00 & 0.00 & 0.89 \\
			
			& Self-Refine
			& 1.00 & 1.00 & 0.83 & 1.00 & 1.00
			& 1.00 & 0.50 & 0.97 & 1.00 & 0.00 & 0.83 \\
			
			& P\&D-Tree
			& 1.00 & 1.00 & 1.00 & 1.00 & 1.00
			& 1.00 & 0.90 & 1.00 & 0.90 & 0.00 & 0.88 \\
			
			& \textbf{LLM-DMGR}
			& \textbf{1.00} & \textbf{1.00} & \textbf{1.00}
			& \textbf{1.00} & \textbf{1.00} & \textbf{1.00}
			& \textbf{1.00} & \textbf{1.00} & \textbf{1.00}
			& \textbf{0.47} & \textbf{0.95} \\
			
			\midrule
			
			%\multirow{6}{*}{\makecell{\footnotesize GPT-5\\\footnotesize mini}}
			\multirow{6}{*}{\makecell{GPT-5\\-mini}}
			& \makecell[l]{Direct LLM\\(Single-pass)}
			& 1.00 & 1.00 & 1.00 & 0.67 & 0.50
			& 0.93 & 1.00 & 0.73 & 0.27 & 0.00 & 0.71 \\
			
			& CoT
			& 1.00 & 0.90 & 1.00 & 1.00 & 1.00
			& 1.00 & 0.47 & 1.00 & 1.00 & 0.00 & 0.84 \\
			
			& CoT-SC
			& 1.00 & 1.00 & 1.00 & 1.00 & 1.00
			& 1.00 & 0.20 & 1.00 & 1.00 & 0.00 & 0.82 \\
			
			& Self-Refine
			& 1.00 & 1.00 & 1.00 & 1.00 & 0.97
			& 1.00 & 0.67 & 1.00 & 1.00 & 0.00 & 0.86 \\
			
			& P\&D-Tree
			& 1.00 & 1.00 & 1.00 & 1.00 & 1.00
			& 1.00 & 1.00 & 0.10 & 0.80 & 0.00 & 0.79 \\
			
			& \textbf{LLM-DMGR}
			& \textbf{1.00} & \textbf{1.00} & \textbf{1.00}
			& \textbf{1.00} & \textbf{1.00} & \textbf{1.00}
			& \textbf{1.00} & \textbf{1.00} & \textbf{1.00}
			& \textbf{0.33} & \textbf{0.93} \\
			
			\bottomrule
		\end{tabular}
	\end{table*}

	\begin{table*}[t]
		\centering
		\renewcommand{\arraystretch}{0.95}
		\setlength{\tabcolsep}{2pt}
		\caption{Coverage of domain preconditions and effects for LLM-DMGR}
		\label{tab:acc_cover}
		
		\begin{tabular}{llccccccccccc}
			\toprule
			\textbf{Model} & \textbf{Type}
			& Gripper
			& Gripper-ood
			& Miconic
			& Movie
			& Driverlog
			& Hiking
			& Termes
			& Floortile
			& Childsnack
			& Barman
			& \textbf{Avg.} \\
			\midrule
			
			\multirow{2}{*}{DeepSeek-v4-flash}
			& Precondition
			& 1.00 & 1.00 & 0.93 & 1.00 & 1.00
			& 0.98 & 1.00 & 1.00 & 0.97 & 0.92
			& \textbf{0.98} \\
			
			& Effect
			& 1.00 & 1.00 & 1.00 & 0.85 & 1.00
			& 1.00 & 1.00 & 1.00 & 1.00 & 0.87
			& \textbf{0.97} \\
			
			\midrule
			
			\multirow{2}{*}{GPT-5-mini}
			& Precondition
			& 1.00 & 1.00 & 1.00 & 1.00 & 1.00
			& 0.98 & 0.98 & 1.00 & 1.00 & 0.93
			& \textbf{0.99} \\
			
			& Effect
			& 0.88 & 1.00 & 1.00 & 0.85 & 1.00
			& 1.00 & 1.00 & 1.00 & 1.00 & 0.92
			& \textbf{0.97} \\
			
			\bottomrule
		\end{tabular}
	\end{table*}

	\begin{table*}[t]
		\centering
		\setlength{\tabcolsep}{2.5pt}
		\renewcommand{\arraystretch}{0.95}
		\caption{Results on ablation study of LLM-DMGR in ten domains}
		\label{tab:abl}
		
		\begin{tabular}{cccccccccccc}
			\toprule
			\textbf{Model}
			& Grippers
			& Grippers-ood
			& Miconic
			& Movie
			& Driverlog
			& Hiking
			& Termes
			& Floortile
			& Childsnack
			& Barman
			& \textbf{Avg.} \\
			\midrule
			
			\makecell{\textbf{LLM-DMGR}\\(DeepSeek-v4-flash)}
			& \textbf{1.00} & \textbf{1.00} & \textbf{1.00} & \textbf{1.00}
			& \textbf{1.00} & \textbf{1.00} & \textbf{1.00} & \textbf{1.00}
			& \textbf{1.00} & \textbf{0.47} & \textbf{0.95} \\
			
			--Extraction
			& 1.00 & 1.00 & 1.00 & 1.00 & 1.00 & 1.00 & 1.00 & 0.90
			& 0.33 & 0.30 & 0.85 \\
			
			--Action-Level
			& 1.00 & 1.00 & 1.00 & 1.00 & 1.00 & 1.00 & 1.00 & 0.87
			& 0.50 & 0.37 & 0.87 \\
			
			--Plan-Level
			& 1.00 & 1.00 & 1.00 & 1.00 & 1.00 & 1.00 & 1.00 & 1.00
			& 0.67 & 0.30 & 0.90 \\
			
			Direct LLM (Multi-round)
			& 1.00 & 1.00 & 0.73 & 1.00 & 1.00 & 0.97 & 0.33 & 0.27
			& 0.33 & 0.00 & 0.66 \\
			
			Direct LLM (Single-pass)
			& 1.00 & 1.00 & 0.73 & 1.00 & 0.10 & 0.83 & 0.33 & 0.27
			& 0.00 & 0.00 & 0.53 \\
			
			\midrule
			
			\makecell{\textbf{LLM-DMGR}\\(GPT-5-mini)}
			& \textbf{1.00} & \textbf{1.00} & \textbf{1.00} & \textbf{1.00}
			& \textbf{1.00} & \textbf{1.00} & \textbf{1.00} & \textbf{1.00}
			& \textbf{1.00} & \textbf{0.33} & \textbf{0.93} \\
			
			--Extraction
			& 1.00 & 1.00 & 1.00 & 0.67 & 1.00 & 0.97 & 1.00 & 0.83
			& 0.97 & 0.20 & 0.86 \\
			
			--Action-Level
			& 1.00 & 1.00 & 1.00 & 1.00 & 1.00 & 1.00 & 1.00 & 0.93
			& 0.93 & 0.20 & 0.91 \\
			
			--Plan-Level
			& 1.00 & 1.00 & 1.00 & 1.00 & 1.00 & 1.00 & 1.00 & 0.93
			& 1.00 & 0.07 & 0.90 \\
			
			Direct LLM (Multi-round)
			& 1.00 & 1.00 & 1.00 & 0.67 & 1.00 & 0.97 & 1.00 & 0.77
			& 0.33 & 0.00 & 0.77 \\
			
			Direct LLM (Single-pass)
			& 1.00 & 1.00 & 1.00 & 0.67 & 0.50 & 0.93 & 1.00 & 0.73
			& 0.27 & 0.00 & 0.71 \\
			
			\bottomrule
		\end{tabular}
	\end{table*}

	Following prior work~\cite{mahdavi2024leveraging}, we use the problem-solving rate as the primary evaluation metric, which measures the fraction of planning problems successfully solved under a generated domain model and the corresponding adapted PDDL problems. Formally, for a domain $\mathcal{D}$ with a set of planning problems $\mathbb{P}_{\mathrm{test}} = \{P_1, \dots, P_N\}$, the problem-solving rate is defined as:
	\begin{equation}
		\text{PSR}(\mathcal{D}_{gen}) = \frac{1}{N} \sum_{i=1}^{N} \mathbb{I}\big(\text{Planner}(\mathcal{D}_{gen}, P_i) \text{ succeeds}\big),
	\end{equation}
	where $N=10$, $\mathbb{I}(\cdot)$ is the indicator function, and $\mathcal{D}_{gen}$ denotes the generated domain model.

	To further assess generated domain models, we report \emph{semantic coverage}: the fraction of ground-truth action preconditions and effects recovered under semantic alignment rather than exact string matching, since generated predicates may differ in surface form (e.g., names or argument ordering) while preserving equivalent meanings.
	Let $\mathcal{C}_{\text{gt}}^{\text{pre}}$ and $\mathcal{C}_{\text{gen}}^{\text{pre}}$ (\ $\mathcal{C}_{\text{gt}}^{\text{eff}}$ and $\mathcal{C}_{\text{gen}}^{\text{eff}}$) denote the sets of precondition (\ effect) literals in the ground-truth and generated domain models.
	Alignment is performed \emph{action-wise}: a ground-truth literal of action $a$ may only match a generated literal of the same action.
	Two literals $c_i$ and $c_j$ are semantically equivalent ($c_i \leftrightarrow c_j$) iff they share the same polarity (precondition, add-effect, or delete-effect), compatible arity and argument types, equivalent participant roles under a consistent argument remapping within the action schema, and the same environmental meaning under the domain description (synonymous predicate names are allowed); partial matches are not counted.
	Alignments are proposed by an LLM given the ground-truth and generated action schemas, then verified by two authors experienced in classical planning and PDDL; scores use only the human-verified alignments.
	This metric evaluates \emph{recall} of ground-truth literals, not precision of generated ones.
	
	We define the semantically aligned precondition and effect sets as:
	\begin{equation}
		\mathcal{M}_{\text{pre}}
		=\left\{
		c_i  \in \mathcal{C}_{\text{gt}}^{\text{pre}} \middle| \exists c_j \in \mathcal{C}_{\text{gen}}^{\text{pre}},	
		c_i \leftrightarrow	c_j
		\right\},
	\end{equation}
	\begin{equation}
		\mathcal{M}_{\text{eff}}
		=\left\{
		c_i  \in \mathcal{C}_{\text{gt}}^{\text{eff}} \middle| \exists c_j \in \mathcal{C}_{\text{gen}}^{\text{eff}},	
		c_i \leftrightarrow	c_j
		\right\},
	\end{equation}
	where $\leftrightarrow$ follows the equivalence criteria above.
	The precondition and effect coverage are computed as:
	\begin{equation}
		\text{Coverage}_{\text{pre}}	=\frac{|\mathcal{M}_{\text{pre}}|}{|\mathcal{C}_{\text{gt}}^{\text{pre}}|},
		\quad
		\text{Coverage}_{\text{eff}}	=\frac{|\mathcal{M}_{\text{eff}}|}{|\mathcal{C}_{\text{gt}}^{\text{eff}}|}.
	\end{equation}

	\subsection{Main Results}
	
	\begin{table}[t]
		\centering
		\setlength{\tabcolsep}{3pt}
		\renewcommand{\arraystretch}{0.95}
		\caption{Total token usage of each method for one complete run over all ten domains.}
		\label{tab:token}
		\begin{tabular}{lcc}
			\toprule
			\textbf{Method}
			& \textbf{DeepSeek-v4-flash}
			& \textbf{GPT-5-mini} \\
			\midrule
			Direct LLM (Single-pass)
			& 1.03M & 1.13M \\
			CoT
			& 1.15M & 1.24M \\
			CoT-SC
			& 1.49M & 1.63M \\
			Self-Refine
			& 1.31M & 1.43M \\
			Mahdavi et al. (P\&D-Tree)
			& 6.63M & 6.12M\\
			LLM-DMGR
			& 2.94M & 2.36M \\
			\bottomrule
		\end{tabular}
	\end{table}

	Table~\ref{tab:main} compares the problem-solving rates of LLM-DMGR with representative matched-input baselines, including CoT, self-consistency CoT, and Self-Refine, together with our same-backbone reproductions of P\&D-Tree~\cite{mahdavi2024leveraging}. Overall, these reasoning and refinement baselines substantially improve over direct LLM generation, but their performance remains sensitive to both the planning domain and the underlying backbone. With DeepSeek-v4-flash, self-consistency CoT achieve the strongest baseline average of 0.89, whereas Self-Refine performs best among the GPT-5-mini baselines with an average of 0.86. Nevertheless, substantial degradation is still observed on challenging domains. In particular, performance on Termes varies considerably across methods and backbones, while all baseline methods fail completely on Barman. P\&D-Tree also exhibits pronounced backbone-dependent variation, dropping to 0.10 on Floortile and 0.80 on Childsnack with GPT-5-mini despite achieving 1.00 and 0.90, respectively, with DeepSeek-v4-flash.
	
	In contrast, LLM-DMGR achieves the highest average problem-solving rates on both backbones, reaching 0.95 with DeepSeek-v4-flash and 0.93 with GPT-5-mini. It obtains perfect problem-solving rates on both backbones for Termes and Floortile, and, is the only evaluated method that solves a non-trivial fraction of Barman instances on both backbones, achieving 0.47 and 0.33, respectively. These results indicate that the gains of LLM-DMGR cannot be explained solely by stronger prompting, self-consistency, iterative refinement, or backbone choice. Instead, they support the effectiveness of explicitly incorporating semantic knowledge and structured refinement mechanisms into the domain-model generation process.
	
	%Table~\ref{tab:acc_cover} evaluates the semantic coverage of generated action preconditions and effects. LLM-DMGR achieves near-complete coverage on both backbones, with average precondition/effect coverage of 0.98/0.97 and 0.99/0.97, respectively. Most remaining errors are concentrated in the more challenging domains, particularly Barman. The remaining gap between high semantic coverage and problem-solving performance also suggests that coverage alone does not guarantee a valid domain model: a small number of planning-critical inconsistencies can still prevent successful planning.

	Table~\ref{tab:acc_cover} evaluates the semantic coverage of generated action preconditions and effects. LLM-DMGR achieves high average coverage on both backbones, reaching 0.98/0.97 for DeepSeek-v4-flash and 0.99/0.97 for GPT-5-mini. While lower coverage is more evident in challenging domains such as Barman, incomplete effect coverage also appears in simpler domains such as Movie and Grippers. Importantly, high coverage alone does not guarantee a valid domain model, as remaining omissions, incorrect literals, or cross-action inconsistencies may still affect planning success.
	
	 Table~\ref{tab:token} further compares the computational cost of the evaluated methods, accounting for all LLM calls across all rounds, including generation, evaluation, multi-agent debate, and feedback synthesis. Prompting-only approaches require the fewest tokens but generally achieve lower problem-solving rates. Among them, Self-Refine consumes more tokens than CoT because it iteratively generates feedback and revises the domain, while remaining below self-consistency CoT. LLM-DMGR introduces additional cost through its bi-level refinement process. Nevertheless, LLM-DMGR requires less than half the token consumption of the P\&D-Tree reproductions while achieving higher overall problem-solving rates. This result suggests that its additional computation is used more effectively through structured action-level and plan-level refinement.

	\subsection{Ablation Studies}

	Table~\ref{tab:abl} presents the ablation results of LLM-DMGR in terms of problem-solving rate. We remove semantic knowledge extraction (\textbf{--Extraction}), action-level refinement (\textbf{--Action-Level}), and plan-level agentic refinement (\textbf{--Plan-Level}) individually while keeping the remaining components unchanged. The \textbf{--Plan-Level} variant removes the entire plan-level refinement module, including planner feedback and multi-agent feedback processing during refinement. We additionally include \textbf{Direct LLM (Multi-round)}, which uses the same backbone and maximum number of refinement iterations as LLM-DMGR but directly regenerates the domain from planner feedback without the proposed semantic extraction and bi-level refinement mechanisms. \textbf{Direct LLM (Single-pass)} performs domain generation without iterative refinement.
	
	The results show that the contribution of each component becomes more evident as domain complexity increases. On relatively simple domains such as Grippers and Grippers-ood, even single-pass generation is generally sufficient to construct usable domain models, leaving limited room for further refinement. In contrast, removing semantic knowledge extraction leads to consistent degradation on more challenging domains such as Floortile, Childsnack, and Barman. This suggests that explicitly extracted semantic knowledge provides useful prior information for constructing action models when domain descriptions involve more complex predicates and action dependencies.
	
	Action-level refinement further improves the reliability of individual action models, with its removal primarily affecting the harder domains. This indicates that local validation and refinement are particularly useful for correcting errors in action preconditions and effects before they propagate to the complete domain model. Plan-level agentic refinement provides a complementary source of global feedback by evaluating the assembled domain through symbolic planning. Its contribution is most evident on Barman, where removing this component causes a substantial decrease on both backbones. This observation suggests that some defects cannot be reliably identified from individual actions alone and require reasoning about their interactions during planning.
	
	%Finally, the comparison with \textbf{Direct LLM (Multi-round)} distinguishes the effect of structured refinement from simply increasing the number of LLM calls. Despite using the same maximum number of refinement iterations, direct regeneration achieves substantially lower average problem-solving rates than LLM-DMGR (0.66 vs.\ 0.91 on DeepSeek-v4-flash and 0.77 vs.\ 0.93 on GPT-5-mini). Therefore, the improvement of LLM-DMGR cannot be explained merely by additional generation rounds; rather, the semantic knowledge extraction and bi-level refinement mechanisms provide structured feedback that progressively improves the reliability of the generated domain models.
	
	Finally, we compare LLM-DMGR with \textbf{Direct LLM (Multi-round)} using the same backbone and maximum number of iterations. LLM-DMGR achieves higher average problem-solving rates: 0.95 vs.\ 0.66 on DeepSeek-v4-flash and 0.93 vs.\ 0.77 on GPT-5-mini. These results show that the complete LLM-DMGR framework outperforms direct regeneration under the same iteration limit.
	
	%%%%%%%%%%%%%%%%%%%%%%%%%%%%%%%%%%%%%%%%%%%%%%%%%%%%%%%%%%%%%%%%%%%%%%%%
	\section{Conclusion}

	In this paper, we presented LLM-DMGR, a semantic-enriched agentic refinement framework for generating PDDL domain models from natural language descriptions. LLM-DMGR integrates structured semantic knowledge extraction with action-level validation and plan-level agentic refinement to progressively improve the reliability and planning usability of generated domain models. Experiments across ten planning domains demonstrate that LLM-DMGR achieves higher problem-solving rates than representative baselines, while ablation studies confirm the effectiveness of its key components and structured refinement strategy. Overall, our results demonstrate the potential of integrating semantic knowledge with agentic refinement for reliable and automated planning domain model generation.
	
	%%%%%%%%%%%%%%%%%%%%%%%%%%%%%%%%%%%%%%%%%%%%%%%%%%%%%%%%%%%%%%%%%%%%%%%%
	
	%%% The acknowledgments section is defined using the "acks" environment
	%%% (rather than an unnumbered section). The use of this environment 
	%%% ensures the proper identification of the section in the article 
	%%% metadata as well as the consistent spelling of the heading.

	%%%%%%%%%%%%%%%%%%%%%%%%%%%%%%%%%%%%%%%%%%%%%%%%%%%%%%%%%%%%%%%%%%%%%%%%
	
	%%% The next two lines define, first, the bibliography style to be 
	%%% applied, and, second, the bibliography file to be used.
	
	\bibliographystyle{ACM-Reference-Format} 
	\bibliography{sample}
	
	%%%%%%%%%%%%%%%%%%%%%%%%%%%%%%%%%%%%%%%%%%%%%%%%%%%%%%%%%%%%%%%%%%%%%%%%
	
\end{document}